\documentclass[sigconf]{acmart}

\usepackage{booktabs} 
\usepackage[ruled,noline]{algorithm2e}
\usepackage{algorithmic}
\usepackage{enumitem, tabularx}
\usepackage{color,soul}
\usepackage{mathtools}
\usepackage{varwidth}
\usepackage{multirow}
\usepackage{diagbox}
\usepackage{subcaption}

\usepackage{enumitem}
\setlist{leftmargin=4mm}

\AtBeginDocument{%
  \providecommand\BibTeX{{%
    \normalfont B\kern-0.5em{\scshape i\kern-0.25em b}\kern-0.8em\TeX}}}

\copyrightyear{2022}
\acmYear{2022}
\setcopyright{acmcopyright}
\acmConference[KDD '22] {Proceedings of the 28th ACM SIGKDD Conference on Knowledge Discovery and Data Mining}{August 14--18, 2022}{Washington, DC, USA.}
\acmBooktitle{Proceedings of the 28th ACM SIGKDD Conference on Knowledge Discovery and Data Mining (KDD '22), August 14--18, 2022, Washington, DC, USA}
\acmPrice{15.00}
\acmISBN{978-1-4503-9385-0/22/08}
\acmDOI{10.1145/3534678.3539419}

\newtheorem{theorem}{Theorem}
\newtheorem{lemma}{Lemma}
\newtheorem{assumption}{Assumption}
\newtheorem{definition}{Definition}

\begin{document}
\setlength{\abovedisplayskip}{2pt}
\setlength{\belowdisplayskip}{2pt}
\setlength{\textfloatsep}{3pt}

\title{RES: A Robust Framework for Guiding Visual Explanation}

\author{Yuyang Gao}
\email{yuyang.gao@emory.edu}
\affiliation{%
 \institution{Emory University}
 \city{Atlanta}
 \state{GA}
 \country{USA}
}
\author{Tong Steven Sun}
\email{tsun8@gmu.edu}
\affiliation{%
 \institution{George Mason University}
 \city{Fairfax}
 \state{VA}
 \country{USA}
}
\author{Guangji Bai}
\email{guangji.bai@emory.edu}
\affiliation{%
 \institution{Emory University}
 \city{Atlanta}
 \state{GA}
 \country{USA}
}
\author{Siyi Gu}
\email{carrie.gu@emory.edu}
\affiliation{%
 \institution{Emory University}
 \city{Atlanta}
 \state{GA}
 \country{USA}
}
\author{Sungsoo Ray Hong}
\email{shong31@gmu.edu}
\affiliation{%
 \institution{George Mason University}
 \city{Fairfax}
 \state{VA}
 \country{USA}
}
\author{Liang Zhao}
\authornote{Corresponding author}
\email{liang.zhao@emory.edu}
\affiliation{%
 \institution{Emory University}
 \city{Atlanta}
 \state{GA}
 \country{USA}
}

\renewcommand{\shortauthors}{Gao, et al.}

\begin{abstract}

Despite the fast progress of explanation techniques in modern Deep Neural Networks (DNNs) where the main focus is handling ``how to generate the explanations'', advanced research questions that examine the quality of the explanation itself (e.g., ``whether the explanations are accurate'') and improve the explanation quality (e.g., ``how to adjust the model to generate more accurate explanations when explanations are inaccurate'') are still relatively under-explored.
To guide the model toward better explanations, techniques in explanation supervision---which add supervision signals on the model explanation---have started to show promising effects on improving both the generalizability as and intrinsic interpretability of Deep Neural Networks.
However, the research on supervising explanations, especially in vision-based applications represented through saliency maps, is in its early stage due to several inherent challenges: 
1) inaccuracy of the human explanation annotation boundary, 2) incompleteness of the human explanation annotation region, and 3) inconsistency of the data distribution between human annotation and model explanation maps.
To address the challenges, we propose a generic RES\footnote{Code available at: \scriptsize \url{https://github.com/YuyangGao/RES}.} framework for guiding visual explanation by developing a novel objective that handles inaccurate boundary, incomplete region, and inconsistent distribution of human annotations, with a theoretical justification on model generalizability.
Extensive experiments on two real-world image datasets demonstrate the effectiveness of the proposed framework on enhancing both the reasonability of the explanation and the performance of the backbone DNNs model. 
\end{abstract}

\ccsdesc[500]{Computing methodologies~Supervised learning}
\ccsdesc[500]{Computing methodologies~Computer vision}
\keywords{Explainability, Interpretability, Robustness, Visual Explanation}


\maketitle
\section{Introduction}

\begin{figure}
    \centering
    \includegraphics[width=0.9\linewidth]{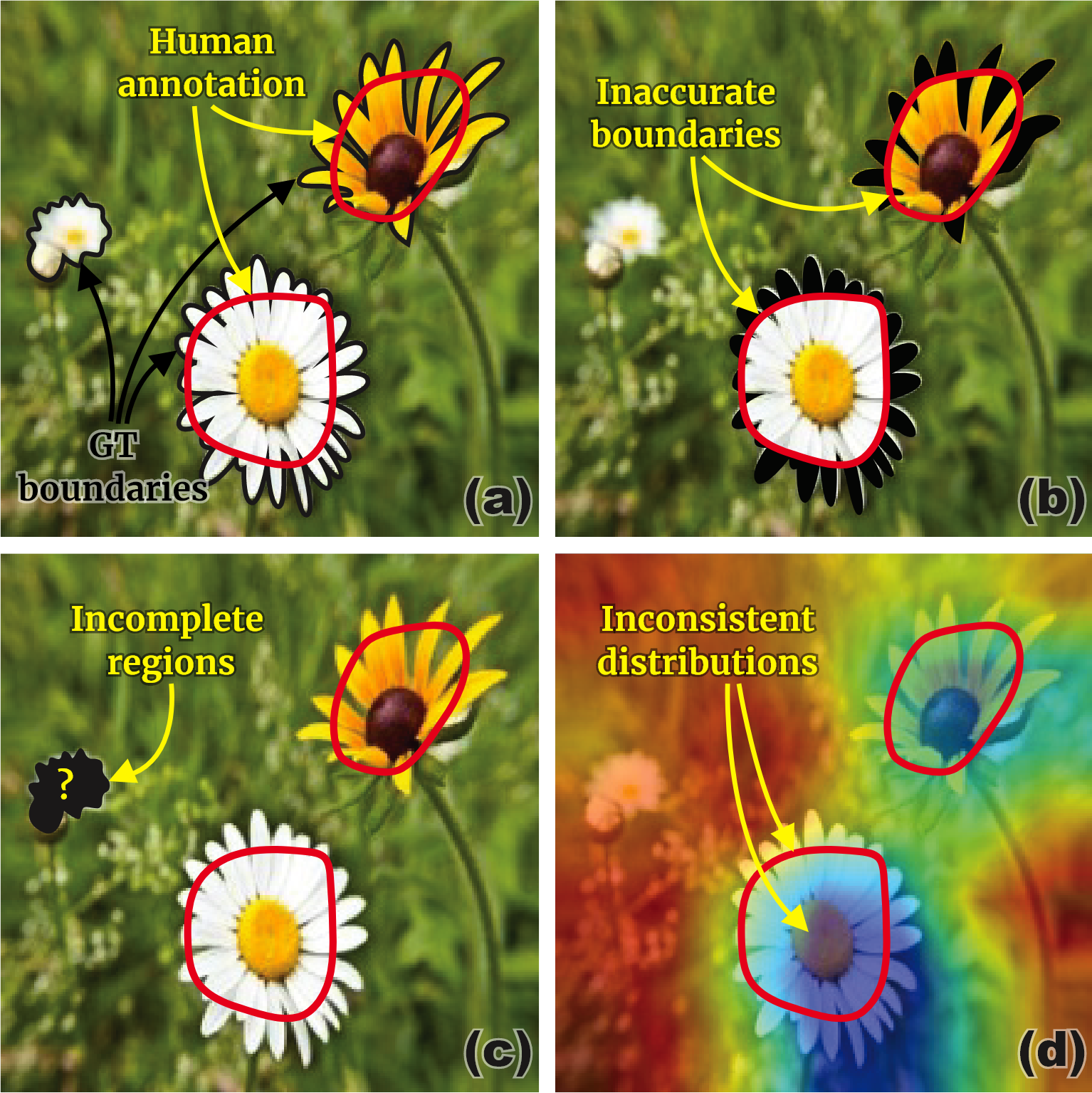}
    \caption{An example showing the challenges present in the human annotation labels: (a) human annotations are represented with red lines while ground-truth boundaries are shown with black lines. (b) Error caused by ``inaccurate boundaries'' are presented with black regions, (c) Error caused by ``incomplete regions'' are shown with a black region, and (d) the discrepancies between the ``binary'' human annotation and the ``continuous'' model-generated explanation maps. The explanation is queried based on predicting the scene as `wild nature'.}
    \label{fig:challenges}
\end{figure}

As DNNs become available in a wide range of application areas, the study on explainability or explainable AI (XAI) is currently attracting considerable attention~\cite{adadi2018peeking, arrieta2020explainable, guidotti2018survey}.
To open the ``black box'' of DNNs, many explainability techniques have been proposed that try to provide the ``local explanation'' of the DNNs prediction for a specific instance~\cite{guidotti2018survey}, such as methods that provide the saliency maps for understanding which sub-parts (i.e., features) in an instance are most responsible for the model prediction~\cite{zhou2016learning, selvaraju2017grad, montavon2019layer, bach2015pixel, baisaliency, montavon2017explaining}. 
While we are witnessing the fast growth of research in local explanation techniques in recent years, the majority of focus is rather handling ``how to generate the explanations'', rather than understanding ``whether the explanations are accurate/reasonable'', ``what if the explanations are inaccurate/unreasonable'', and ``how to adjust the model to generate more accurate/reasonable explanations''. 

Recently, techniques in \textit{explanation supervision}, which support machine learning builders to improve their models by using supervision signals derived from explanation techniques, have started to show promising effects.
The effects include improving both the generalizability and intrinsic interpretability of DNNs in many data types where the human annotation labels can be assigned accurately on each feature of the data.
Such data type includes text data~\cite{jacovi2020aligning, ross2017right} and attributed data~\cite{visotsky2019few}.
However, the research on supervising explanations on image data---where the explanation is represented through saliency maps---is still under-explored~\cite{hong2020human}. In part, this is due to several inherent challenges in supervising visual explanations: 
\textbf{1) Inaccuracy of the human explanation annotation boundary.} It is difficult and costly for humans to make a perfectly accurate boundary which could lead the model to falsely assign positive explanation value to irrelevant features (i.e., pixels in image data). For example, as shown by the yellow arrows in Figure \ref{fig:challenges} (b), the coarsely drawn boundary falsely excluded a non-trivial region of the boundary of the wildflowers that could also be important to the prediction.
\textbf{2) Incompleteness of the human explanation annotation region.} When labeling the explanation for image data, people usually tend to provide only a few regions as long as they are sufficient to convince people about the decision and do not bother to comprehensively find all the possible regions. Such incompleteness can mislead the model to wrongly penalize all the regions as long as they are not selected by annotators. Figure \ref{fig:challenges} (c) shows an example where the human annotation clearly missed one wildflower as shown in the black region.
\textbf{3) Inconsistency of the data distribution between human annotation and model visual explanations.}The saliency maps generated by model explainers are continuous (e.g., Fig. \ref{fig:challenges} (d), heatmap) whereas human annotations are typically binary 'e.g., red circled areas annotated from humans in Fig. \ref{fig:challenges} (d) represent positive while the rest of areas are negative).
Therefore, human-annotated explanations cannot be directly used to supervise the model and its explanations without significant efforts to fill the gap between the data domain and distributions.

To address the above challenges, beyond merely applying human annotation labels directly as the supervision signals to train the model, this work focuses on proposing a generic robust explanation supervision framework for learning to explain DNNs under the assumptions that the human annotation labels can be inaccurate in the boundary, incomplete in the region, as well as inconsistent with the distribution of the model explanation.
Specifically, we propose a novel robust explanation loss that addresses all three aforementioned challenges present in the human annotation labels that can be noisy~\cite{chung2021understanding, chung2019efficient}. 
In addition, we give a theoretical justification of the benefits of having the proposed explanation loss to the generalizability power of the backbone DNN model.

Specifically, the main contributions of our study are as follows:
\begin{enumerate}
 \item \textbf{Proposing a generic framework for learning to explain DNNs with explanation supervision.} 
 We propose a unified framework that enables explanation supervision on DNNs with both positive and negative explanation annotation labels and is generalizable to the existing differentiable explanation methods.

 \item \textbf{Developing a robust model objective that can handle the noisy human annotation labels as the supervision signal.}
 We propose a novel robust explanation loss that can handle the inaccurate boundary, incomplete region, as well as inconsistent distribution challenges in applying the noisy human annotation labels as the supervision signal.

 \item \textbf{Providing a theoretical justification on the generalizability power of the proposed framework.} 
 We formally derive a theorem that provides an upper bound for the generalization error of applying the proposed robust explanation loss when training the backbone DNN models.
 
 \item \textbf{Conducting comprehensive quantitative and qualitative experimental analysis to validate the effectiveness of the proposed model.} 
 Extensive experiments on two real-world image datasets, gender classification and scene recognition, demonstrate that the proposed framework improved the backbone DNNs both in terms of prediction power and explainability. In addition, qualitative analyses, including case studies and user studies of the model explanation, are provided to demonstrate the effectiveness of the proposed framework.
\end{enumerate}

\section{Related work}
Our work draws inspiration from the research fields of local explainability techniques of DNNs that provide the model-generated explanation, and explanation supervision on DNNs which enables the design of pipelines for the human-in-the-loop adjustment on the DNNs based on their explanations to enhance both explainability and performance of DNN models.

\subsection{Local Explainability Techniques of DNNs}
As DNNs become widely deployed in a wide spectrum of application areas, recent years have seen an explosion of research in understanding how DNNs work under the hood (e.g., explainable AI, or XAI)~\cite{arrieta2020explainable, guidotti2018survey, gao2021bean, yu2019interpreting, hong2020human}.
Due to the ``black box'' nature of DNNs, most of the existing and well-received explainability methods focus on providing a ``local explanation'' that aims at explaining the prediction in understandable terms for humans for a specific instance or record~\cite{guidotti2018survey}. 
One popular direction is to compute saliency maps as the local explanation, which provide the saliency values regarding which input features are most responsible for the prediction of the model~\cite{zhou2016learning, selvaraju2017grad, montavon2019layer, bach2015pixel, montavon2017explaining}. 
For example, for image input, a saliency map is able to summarize where the model is ``paying attention to'' when performing a certain image recognition task. 
In this direction, one set of works incorporates network activations into their visualizations, such as Class Activation Mapping (CAM)~\cite{zhou2016learning} and Grad-CAM~\cite{selvaraju2017grad}.
Another set of approaches takes a backward pass and assigns a relevance score for each layer backpropagating the effect of a decision up to the input level, existing works such as LRP~\cite{montavon2019layer, bach2015pixel}, and DTD~\cite{montavon2017explaining} belong to this category.
In addition, some model inspection methods such as VisualBackProp (VBP)~\cite{bojarski2016visualbackprop} can also provide a local explanation similar to the LRP approaches. 
Besides the above techniques that are more specifically designed for interpreting image data, there are also several existing techniques that aim at providing more model-agnostic explanations on different types of data, such as LIME~\cite{ribeiro2016should} and Anchors~\cite{ribeiro2018anchors}.
Please refer to the survey papers \cite{arrieta2020explainable, guidotti2018survey} for a more comprehensive review of the existing works.

\subsection{Explanation Supervision on DNNs}
The potential of using explanation--methods devised for understanding which sub-parts in an instance are important for making a prediction--in improving DNNs has been studied in many domains across different applications~\cite{gao2021gnes}.
In particular, explanation supervision techniques have been widely explored on image data by the computer vision community \cite{linsley2018learning, mitsuhara2019embedding, patro2020explanation, zhang2019interpretable, das2017human}.
Existing studies have shown the benefit of using stronger supervisory signals by teaching networks where to attend~\cite{linsley2018learning}.
Following this line of study, several explanation supervision frameworks have been proposed. 
Mitsuhara et al. \cite{mitsuhara2019embedding} proposed a post hoc fine-tuning strategy, where an end-user is asked to manually edit the model's explanation to interactively adjust its output. However, the proposed framework is only applicable to a specific type of DNN called Attention Branch Network~\cite{fukui2019attention}.
In addition, several frameworks designed for the Visual Question Answering (VQA) domain have been proposed, where the goal is to obtain the improved explanation on both the text data and the image data \cite{ zhang2019interpretable, patro2020explanation, das2017human}.

Recently, several more generic frameworks have been proposed for explanation supervision on image data.
One existing work proposed a conceptual framework HAICS~\cite{shen2021human}, and the authors further implement it in an image classification application with human annotation in the form of scribble annotations as explanation supervision signals.
Another noteworthy work has proposed the Interactive Attention Mechanism~\cite{gao2022aligning} which helps humans to spot cases with unreasonable local explanation and directly adjust it using GRADIA.
Using the adjusted feedback from human users, GRADIA aims at improving the performance and quality of explanation.
Besides image data, the explanation supervision has also been studied on other data types, such as texts~\cite{jacovi2020aligning, ross2017right, choi2019AILA}, attributed data~\cite{visotsky2019few}, and more recently on graph-structured data~\cite{gao2021gnes}.
However, most of the existing works typically assume the human labels are clean and accurate, while in practice they are prone to be inexact, inaccurate, and incomplete when directly used as the supervision signal for supervising the model explanation. To our best knowledge, we are the first to propose a robust explanation supervision framework that aims at handling this open research problem.

\section{Model}
In this section, we first introduce the proposed RES framework that enables explanation supervision on DNNs with both positive and negative explanation annotation labels. 
We then move on to propose a novel robust explanation loss that is designed to handle the inaccurate boundary, incomplete region, as well as inconsistent distribution challenges in applying the noisy human annotation labels as the supervision signal.
Finally, we give the theoretical justification of the benefits of having the proposed explanation loss to the generalizability power of the backbone DNN model.

\textbf{Problem formulation:}
Let $x\in\mathbb{R}^{C \times H \times W}$ be the input image data with $C$ channels, $H$ as height, and $W$ as width. Let $y$ be the class label for input $x$, the general goal for a DNN model is to learn the mapping function $f$ for each input $x$ to its corresponding label, $f: x \rightarrow y$.

\subsection{The RES Framework}
The general goal for the RES framework is to boost the model explainability via robust explanation supervision such that the model can robustly learn to assign more importance to the right input features even given noisy human explanation annotation labels, and consequently boost the task performance as well as the interpretability of the backbone DNN model. 
Here, we present the general learning objective of the RES framework to be a joint optimization of the model prediction loss and the robust explanation loss.
Concretely, we propose the objective function as:
\begin{equation}
\label{eq:overall}
\min \ 
\sum^N_i
\underbrace{
\mathcal{L}_{\text{Pred}}(f(x^{(i)}),y^{(i)})
}_{\mbox{\small prediction loss}}
+ 
\underbrace{
\mathcal{L}_{\text{Exp}} (\langle M^{(i)}, F^{(i)}, C^{(i)} \rangle)
}_{\mbox{\small robust explanation loss}}
\end{equation}
where $M^{(i)} \in\mathbb{R}^{H \times W}$ denotes the model-generated explanations for $i$th sample using a given explanation method; $F^{(i)}\in\{0, 1\}^{H \times W}$ and $C^{(i)}\in\{0, 1\}^{H \times W}$ denote the corresponding binary labels for positive (i.e., $F^{(i)}_{j,k}=1$ if the pixel at coordinate $(j,k)$ of sample image $i$ should be assigned with high importance, and $0$ otherwise) and negative (i.e., $C^{(i)}_{j,k}=1$ if the pixel at coordinate $(j,k)$ of image $i$ should be assigned with low importance value, and $0$ otherwise) explanation marked by the human annotators.
$\mathcal{L}_{\text{Pred}}(f(x^{(i)}),y^{(i)})$ is the typical prediction loss (such as the cross-entropy loss). 

\subsection{Robust Explanation Supervision for Noisy Explanation Annotation labels}
To address the challenges presented in the noisy human annotation labels, we propose a robust explanation loss $\mathcal{L}_{\text{Exp}}$ that measures the discrepancies between model and human explanations regarding both the positive and negative explanation and taking into consideration the noisy nature of human annotation labels. Without loss of generality,
let us assume $\tilde{M^{(i)}} = \tilde{F}^{(i)}-\tilde{C}^{(i)}$ in range $[-1, 1]$ be the ground truth ideal explanation value for input image $x^{(i)}$, given the ideal positive explanation $\tilde{F}^{(i)}\in[0,1]$ and negative explanation $\tilde{C}^{(i)}\in[0,1]$; the binary human annotation as $F^{(i)}$ and $C^{(i)}$; and the model explanation as $M^{(i)}=g(f_{\theta}((x^{(i)}))$, where function $g(\cdot)$ specify the explanation method. We have $\mathbb{E}[\|M^{(i)}-(F^{(i)}-C^{(i)})\|-\|(F^{(i)}-C^{(i)})-\tilde{M}^{(i)}\|]\le \max\{0,\mathbb{E}[\|M^{(i)}-(F^{(i)}-C^{(i)})\|]-\mathbb{E}[\|(F^{(i)}-C^{(i)})-\tilde{M}^{(i)}\|]\}\le  \mathbb{E}[\max\{0,\|M^{(i)}-(F^{(i)}-C^{(i)})\|-\|(F^{(i)}-C^{(i)})-\tilde{M}^{(i)}\|\}] \le \mathbb{E}[\|M^{(i)}-\tilde{M}^{(i)}\|]$ according triangle inequality. We define $\alpha = \mathbb{E}[\|(F^{(i)}-\tilde{F}^{(i)}) - (C^{(i)}-\tilde{C}^{(i)})\|]$.
Therefore, to minimize $\|M^{(i)}-\tilde{M}^{(i)}\|$, we can have a tighter surrogate loss based on the annotated labels as follows:
\begin{gather}
    \max\{0,\|M^{(i)}-(\tilde{F}^{(i)}-\tilde{C}^{(i)})\|-\alpha\} \notag
\end{gather}

Since the ground truth $\tilde{F}$ and $\tilde{C}$ are unknown, estimating $\alpha$ can be difficult. In practice, we can assume their distributions are positively correlated with the distribution of $F$ and $C$, which can therefore be estimated by a slack variable $\alpha$. To keep it simple and without loss of generality, in this work, we define $\alpha$ as a hyper-parameter of the framework assuming no additional knowledge about the ideal distribution. 

\subsubsection{Bridging the distribution between human labels and model explanation maps}
To bridge the continuous model explanation $M^{(i)}$ with binary human labels $C$ and $F$, we propose to split the above objective into two terms with bidirectional projections, as:
\begin{gather}
    \min_{\theta, a}
    \sum\nolimits^N_i
    \max\{0,\|[\hat{M}^{(i)}-(F^{(i)}-C^{(i)})]\|-\alpha\} \notag \\
    +d(M^{(i)},h(F^{(i)},C^{(i)}))
\end{gather}
where $d(\cdot)$ is a distance function, $h(\cdot)$ is a mapping function that maps the binary masks $F^{(i)}$ and $C^{(i)}$ to continuous value in range $[0,1]$, and $\hat{M}^{(i)}$ is a binary projection of $M^{(i)}$ by a threshold $a$, as:
\begin{equation}
\label{eq:M_hat}
\hat{M}^{(i)} = \left\{
        \begin{array}{ll}
            1  & \quad M^{(i)}\geq a \\
            -1 & \quad M^{(i)}<a \\
        \end{array}
    \right.
\end{equation}

Basically, the above equation takes both the absolute difference (measured by the first term) and relative distance (measured by the second term) into consideration when comparing the continuous model explanation and the binary human explanation masks. 

\subsubsection{Mitigating the Inaccurate Boundary via Label Imputation}
To realize the mapping function $h(\cdot)$ in Equation \eqref{eq:M_hat} which aims at projecting the binary human labels into continuous value domain, an intuitive way is to define $h(\cdot)$ as applying a $k \times k$ Gaussian kernel on the binary annotation labels $F$ and $C$ such that the pixels that close to the boundary of the manual label will also obtain slack values to boost the robustness and deal with the inexact and inaccurate boundary from human annotation. 

However, a pre-defined kernel matrix might not be suitable for every data sample, and the discrepancy and inconsistency among annotators can also influence the accuracy of such a pre-defined estimation on handling the inaccurate boundary issue. 
Therefore, we further extend this idea and define a learnable imputation function $h_\phi(\cdot)$ with multiple learnable kernel transformations as the parameter set $\phi$, such that the kernels' weights can be adjusted and learned to make better estimations of the ground truth explanation values and provide better mitigation to the inaccurate boundary problem. 
Specifically, the explanation loss with a learnable imputation function is as follows:
\begin{gather}
    \min_{\theta, a, \phi}
    \sum\nolimits^N_i
    \max\{0,\|[\hat{M}^{(i)}-(F^{(i)}-C^{(i)})]\|-\alpha\} \notag\\
    +d(M^{(i)},h_\phi(F^{(i)},C^{(i)}))
    \label{eq:imp}
\end{gather}
where $\phi$ is the parameter set of the imputation function $h_\phi(\cdot)$. The imputation function can be realized by applying multiple layers of convolution operations with learnable kernels over the raw annotation label $F$ and $C$.

\subsubsection{Handling the Incomplete Region by Selective Penalization}
Finally, due to the incompleteness of human annotation labels, and to avoid falsely penalizing the model from assigning importance to the relevant features missed by the human labels, we propose to only selectively apply the explanation supervision signal onto the features with either positive or negative annotation labels.
Concretely, we define the robust explanation loss $\mathcal{L}_{Exp}$ as follows:
\begin{gather}
    \min_{\theta, a, \phi}
    \sum\nolimits^N_i
    \max\{0,\|[\hat{M}^{(i)}-(F^{(i)}-C^{(i)})]\cdot \mathbf{1}(F^{(i)}-C^{(i)}\neq 0)\|-\alpha\} \notag\\
    +d(M^{(i)}\!\cdot \!\mathbf{1}(F^{(i)}\!-\!C^{(i)}\neq 0),h_\phi(F^{(i)},C^{(i)})\!\cdot\! \mathbf{1}(F^{(i)}\!-\!C^{(i)}\neq 0))
    \label{eq:exp_loss}
\end{gather}
where $\mathbf{1}(\cdot)$ is the indicator function, and $\cdot$ represents the elemental-wise multiplication operation.
This formulation also gives the model a certain degree of flexibility on deciding the importance of unlabeled features based on data and downstream task, thus could yield a more generalizable and reasonable explanation that enhance both explainability as well as task performance of the model.

\subsection{Optimization of Robust Explanation Loss}
The indicator function for calculating $\hat{M}^{(i)}$ (as shown in Equation \eqref{eq:M_hat}) prevents us from directly optimizing our model objective with conventional gradient descent algorithms such as Adam~\cite{kingma2014adam}. 
Concretely, the optimization problem presented in Equation \eqref{eq:exp_loss} involves optimizing both the adaptive threshold $a$ and the model-generated explanation $M^{(i)}=g(f_\theta(x^{(i)}))$. 
Here, we propose to first find the optimal threshold $a$ given model parameter $\theta$, and then optimize $\theta$ with a conventional gradient descent algorithm by proposing a differentiable approximation to the indicator function.

First, to find the optimal $a$ given $\theta$, we need to solve the following objective:
\begin{align}
    \min_{a} \sum\nolimits^N_i \|[\hat{M}^{(i)}-(F^{(i)}-C^{(i)})]\cdot \mathbf{1}(F^{(i)}-C^{(i)}\neq 0)\|
\end{align}
Which is equivalent to the following by expanding $\hat{M}^{(i)}$:
\begin{gather}
    \min_{a} \sum\nolimits^N_i \|[\mathbf{1}(M^{(i)}\geq a)-F^{(i)}]\cdot F^{(i)}\|\notag\\ 
    +\|[\mathbf{1}(M^{(i)}<a)-C^{(i)}]\cdot C^{(i)}\|
\end{gather}
If we treat each entry of $M^{(i)}$ as having two inequality constraints on $a$, we can efficiently solve the above formula in $O(m \log m)$ by our proposed algorithm by treating this optimization problem as finding a $a$ that satisfies the maximum number of inequality constraints, where $m=max(|F|, |C|)$. The details of the proposed searching algorithm can be found in Appendix \ref{ap:a}.

To further enable gradient calculation of $M^{(i)}$ in Equation \eqref{eq:exp_loss}, we propose a surrogate loss using the hyperbolic tangent function $tanh(\cdot)$ to approximate the indicator function, as follows:
\begin{gather}
\min_{\theta, a, \phi} \ 
\sum\nolimits^N_i \max\{0,\|[tanh(\gamma(M^{(i)}-a) )\!-\!H^{(i)}] \cdot  \mathbf{1}(H^{(i)}\neq 0) \|\!-\!\alpha\} \notag\\
+d(M^{(i)}\cdot \mathbf{1}(H^{(i)}\neq 0),h_\phi(F^{(i)},C^{(i)})\cdot \mathbf{1}(H^{(i)}\neq 0))
\label{eq:tanh}
\end{gather}
where $H^{(i)} = F^{(i)}-C^{(i)}$; $\gamma$ controls the slop of the hyperbolic tangent function. Moreover, when $\gamma \rightarrow \infty$ , we can ensure such a approximation can be mathematically equivalent to the original indicator function in Equation \eqref{eq:imp} as shown in the following lemma. 
\begin{lemma} Equation \eqref{eq:tanh} is mathematically equivalent to Equation \eqref{eq:exp_loss} when $\gamma \rightarrow \infty$.
\end{lemma}
\begin{proof}
Please refer to Appendix \ref{ap:lem1} for the proof.
\end{proof}
\subsection{Theoretical Analysis of Generalizablity}
In this subsection, we theoretically justify the generalizability power of the proposed explanation loss, as shown in Theorem \ref{th:1} below.

We consider the regularized expected loss:
\begin{equation}
    \mathcal{L}(f_{\theta}) = \mathbb{E}\left[\mathcal{L}_{\text{Pred}}(f_{\theta}(x),y)+\mathcal{L}_{\text{Exp}}(\nabla f_{\theta}(x))\right]
\label{eq:loss function with regularization}
\end{equation}
where $f_{\theta}$ is any learnable function with parameter $\theta\in\Theta$. In addition, denote the empirical loss as
\begin{equation}
    \hat{\mathcal{L}}(f_{\theta}) = \frac{1}{N}\sum\nolimits_{i=1}^{N}\left(\mathcal{L}_{\text{Pred}}(f_{\theta}(x^{(i)}),y^{(i)})+\mathcal{L}_{\text{Exp}}(\nabla f_{\theta}(x^{(i)}))\right)
\label{eq:empirical loss}
\end{equation}
where $N$ denotes the training sample size. $\nabla f_{\theta}(x)$ denotes the gradient of $f_{\theta}$ on input $x$, which can be used to generate any explanation. We omit the label (namely, $F^{(i)}$ and $C^{(i)}$) in $\mathcal{L}_{\text{Exp}}$ here for more compact notation. Also, we assume that $\mathcal{L}_{\text{Pred}}$ is $L_1$-Lipschitz and $\mathcal{L}_{\text{Exp}}$ is $L_2$-Lipschitz continuous w.r.t its first input, respectively.

\begin{definition}[$\delta$-minimizer] 
A function $f_{\hat{\theta}}$ is said to be a $\delta$-minimizer of $\mathcal{L}(\cdot)$ if
\begin{equation}
    \mathcal{L}(f_{\hat{\theta}}) \leq \inf_{\theta\in\Theta}\mathcal{L}(f_{\theta})+\delta
\label{eq:delta minimizer}
\end{equation}
\end{definition}

\begin{assumption}
\label{ass:1}
Let $f_{\theta^{*}}$ be the solution to Eq.~\eqref{eq:loss function with regularization}. There exists a neural network $f_{\tau}$ with $\tau\in\Theta$ such that
\begin{equation}
    \Vert f_{\tau} - f_{\theta^{*}}\Vert^2 \coloneqq \mathbb{E}\left[\vert f_{\tau} - f_{\theta^{*}}\vert^2 + \vert \nabla f_{\tau} - \nabla f_{\theta^{*}}\vert^2\right] \leq C_{1}^2 \frac{\Vert \theta^{*}\Vert^2}{m^{\gamma}}
\end{equation}
where $C_1$ is some constant, $m$ is a constant related to the number of parameters in $f$, and $\gamma$ is a constant order.
\end{assumption}

\begin{assumption}
\label{ass:2}
Given any neural network $f_{\theta}$ from $\theta\in\Theta$ and i.i.d sample $\{x^{(i)}\}_{i=1}^N$. Given any $0<\epsilon<1$, we assume that
\begin{equation}
  \sup_{\theta\in\Theta} \vert \mathcal{L}(f_{\theta}) - \hat{\mathcal{L}}(f_{\theta})\vert \leq \frac{C_2 (V,m,\epsilon)}{\sqrt{N}}
\end{equation}
with probability at least $1-\epsilon$. $C_2$ relies on set $\Theta$, $m$ and $\epsilon$.
\end{assumption}
\noindent Such an inequality can be ontained using some statistical learning theories like Rademacher complexity.

Now we provide our generalization error bound as follow:
\begin{theorem}[Generalizability of Equation \eqref{eq:overall}]
\label{th:1}
Let $f_{\theta^*}$ be the minimizer of $\mathcal{L}(\cdot)$, $f_{\hat{\theta}}$ be a $\delta$-minimizer of $\hat{\mathcal{L}}$, then given $0<\epsilon<1$, with probability at least $1-\epsilon$ over the choiec of $x^{(i)}$, we have
\begin{equation}
    0\leq \mathcal{L}(f_{\hat{\theta}}) - \mathcal{L}(f_{\theta^{*}}) \leq (L_1 + L_2) \frac{C_1\Vert \theta^{*}\Vert}{m^{\gamma/2}} + \frac{2C_2 (V,m,\epsilon)}{\sqrt{N}} + 2\delta
\label{eq:generalization bound}
\end{equation}
\end{theorem}
\begin{proof}
Please refer to Appendix \ref{ap:th1} for the formal proof.
\end{proof}

Our~\autoref{th:1} provides an upper bound for the generalization error between the numerical optimal solution $\hat{\theta}$ and the theoretical optimal solution $\theta^*$. The first term in the bound corresponds to the approximation error given in the first assumption, the second term corresponds to the quadrature error given in the second assumption, and the last term corresponds to the training error. To reduce the generalization error, we need to increase both the number of parameters and training samples. Meanwhile, the empirical loss is needed to be solved sufficiently well.

\section{Experiments}
We test our RES framework on two application domains, gender classification and scene recognition. We first describe the detailed settings for the experiments and then present the quantitative studies on both model prediction as well as the explanation. 
In addition, we include several qualitative studies, including case studies and user studies, to make a better qualitative assessment of how the proposed model has enhanced the explainability of the backbone DNN models.

\subsection{Experimental Settings}

\textbf{Gender Classification Dataset}: The gender classification\footnote{We are aware that using a binary classification in gender does not reflect on the diverse viewpoint of gender in the real world, and we emphasize that the binary ``gender classification'' task here does not represent our viewpoint on gender.} is one of the widely used tasks in the research of fairness in broader machine learning communities~\cite{zhao2017men, barlas2021see, hendricks2018women}.
We constructed the dataset from the Microsoft COCO dataset\footnote{Available online at: https://cocodataset.org/} \cite{lin2014microsoft} by extracting images that had the word ``men'' or ``women'' in their captions. We then filtered out instances that 1) contain both words, 2) include more than two people, or 3) humans appear in the figure is nearly not recognizable from human eyes. We collected a total of 1,600 images that satisfied our criterion and obtained the human annotation labels for all the image samples with our human annotation UI (please refer to Appendix \ref{ap:UI} for more details). 
For data splitting, we only randomly sampled 100 samples out of the 1,600 images as the training set to better simulate a more practical situation where we only have limited assess to the human explanation labels. 
The rest of the 1,500 data samples were then evenly split as the validation set and test set.

\textbf{Scene Recognition Dataset}: 
We obtained the scene images from the Places365 dataset\footnote{Available online at: http://places2.csail.mit.edu/index.html} \cite{zhou2017places}. The original dataset contains more than 10 million images comprising 400+ unique scene categories. 
Following the macro-class defined by \cite{zhou2017places}, we constructed a binary scene recognition task: nature vs. urban. The data samples for the two classes were randomly sampled from a set of pre-defined categories under macro-class ``nature'' and ``urban'', respectively. 
Specifically, the categories we used to sample the data are listed below:
\begin{itemize}[leftmargin=15pt]
    \item \textit{Nature}: mountain, pond, waterfall, field wild, forest broadleaf, rainforest
    \item \textit{Urban}: house, bridge, campus, tower, street, driveway
\end{itemize}
Notice that the categories are non-comprehensive and the generated datasets are just for the purpose of studying the quality of model explanation. 
We balanced the sample size for each category and collected a total of 1,600 images. Again, we obtained the human annotation labels for all the samples with the human annotation UI, and split the data randomly with sample sizes of 100/750/750 for training, validation, and testing.

\textbf{Evaluation Metrics}:
We evaluate the model in terms of task performance as well as in terms of explainability.
For model performance, we use the conventional prediction accuracy to measure the prediction power of the backbone DNN models as the datasets studied are well imbalanced.
For explainability assessment, we leverage the human-labeled explanation on the test set to assess the quality of the model explanation.
Specifically, we use the Intersection over Union (IoU) score~\cite{bau2017network}, which is calculated by taking the bit-wise intersection and union operations between the ground truth explanation and the binarized model explanation to measure how well the two explanation masks overlap. 
In addition, since the IoU score only assesses the quality of positive explanation, we further compute the precision, recall, and F1-score as additional metrics which provide a more comprehensive evaluation of the model-generated explanation by considering the alignment of both positive and negative explanation.

\textbf{Comparison methods}:
We compare the performance of the RES framework with the vanilla backbone model as the baseline as well as two existing explanation supervision methods, GRAIDA~\cite{gao2022aligning} and HAICS~\cite{shen2021human}.
For the proposed framework, we show two variations: RES-G and RES-L, with different implementations of the imputation function. 
Concretely, we studied the following methods:
\begin{itemize}[leftmargin=15pt]
    \item \textbf{Baseline}: The conventional DNN model that is trained with only the prediction loss.
    
    \item \textbf{GRADIA}~\cite{gao2022aligning}: A framework that trains the DNN model with both the prediction loss as well as a conventional L1 loss that directly minimizes the distance between the continuous model explanation and the binary positive explanation labels.
    
    \item \textbf{HAICS}~\cite{shen2021human}: A framework that trains the DNN model with both the prediction loss as well as a conventional Binary Cross-Entropy (BCE) loss that directly minimizes the distance between the continuous model explanation and the combination of positive and negative binary explanation labels.
    
    \item \textbf{RES-G}: The proposed RES framework with the imputation function $g(\cdot)$ as a fixed value Gaussian convolution filter.
    
    \item \textbf{RES-L}:The proposed RES framework with the learnable imputation function $g_\phi(\cdot)$ via multiple layers of learnable kernels.
\end{itemize}

\textbf{Implementation Details}:
For all the methods studied in this work, the backbone DNN model is based on the pre-trained ResNet50 architecture~\cite{he2016deep}. 
All models were trained for 50 epochs using the ADAM optimizer \cite{kingma2014adam} with a learning rate of 0.0001.
To make a fair comparison on explainability, the model explanations were all generated by the well-recognized explanation technique GradCAM~\cite{selvaraju2017grad}, although other local explanation techniques can also be applied in our framework.
The generated explanation maps are normalized in the range of $(0, 1]$ by dividing the maximum saliency value on each sample for model training as well as visualization. 
When calculating the explanation evaluation metrics, the explanation maps were further binarized by a fixed threshold of $0.5$. 
The hyper-parameter $\alpha$ of the proposed RES framework was set to 0.001 for the gender classification task, and 0.01 for the scene recognition task, based on grid research via prediction accuracy on the validation set.
The detailed implementation of the imputation layers for RES-L can be found in the Appendix \ref{ap:imp}.

\subsection{Performance}

\begin{table*}
  \caption{The performance and model-generated explanation evaluation among the proposed models and the comparison methods on both gender classification and scenes recognition tasks. The results are obtained from 5 individual runs for every setting. The best results for each task are highlighted with boldface font and the second bests are underlined.}
  \centering
  \label{tab:results}
  \begin{tabular}{c|c|c|cccc}
    \toprule
    Dataset & Model   & Accuracy  & IoU    &  Precision     &  Recall      &  F1    \\
    \hline
    \multirow{5}{*}{Gender Classification}
        & Baseline & 68.35 $\pm$ 1.00 & 13.68 $\pm$ 0.89 & 52.68 $\pm$ 0.61 & 56.34 $\pm$ 1.63 & 47.77 $\pm$ 1.14
        \\
        & GRADIA & 70.01 $\pm$ 1.47 & 16.66 $\pm$ 1.10 & 64.07 $\pm$ 2.07 & 51.84 $\pm$ 3.55 & 53.35 $\pm$ 3.08
        \\
        & HAICS & 69.29 $\pm$ 0.50 & 17.56 $\pm$ 0.79 & 60.06 $\pm$ 2.17 & 56.48 $\pm$ 2.13 & 54.90 $\pm$ 2.14
        \\
        & RES-G & \textbf{71.33 $\pm$ 0.53} & \underline{22.97 $\pm$ 0.44} & \textbf{76.47 $\pm$ 0.45} & \underline{63.90 $\pm$ 3.64} & \underline{63.54 $\pm$ 2.29}
        \\
        & RES-L & \underline{70.39 $\pm$ 0.35} & \textbf{23.60 $\pm$ 0.36} & \underline{76.32 $\pm$ 0.77} & \textbf{65.75 $\pm$ 1.20} & \textbf{65.24 $\pm$ 0.74}
        \\
    \hline
    \multirow{5}{*}{Scene Recognition}
        & Baseline & 93.42 $\pm$ 0.43 & 38.55 $\pm$ 0.22 & \textbf{89.67 $\pm$ 0.07} & 60.96 $\pm$ 0.56 & 68.47 $\pm$ 0.46
        \\
        & GRADIA & 95.03  $\pm$ 0.35 & 39.60 $\pm$ 1.13 & 87.98 $\pm$ 0.19 & 63.47 $\pm$ 2.24 & 70.80 $\pm$ 1.84
        \\
        & HAICS & 94.89  $\pm$ 0.20 & 41.29 $\pm$ 0.91 & \underline{88.47 $\pm$ 0.53} & 66.23 $\pm$ 1.00 & 72.95 $\pm$ 0.87
        \\
        & RES-G & \textbf{95.91 $\pm$ 0.31} & \textbf{45.97 $\pm$ 0.12} & 87.54 $\pm$ 0.30 & \underline{82.88 $\pm$ 1.14} & \underline{82.90 $\pm$ 0.33}
        \\
        & RES-L & \underline{95.53 $\pm$ 0.54} & \underline{44.64 $\pm$ 0.31} & 86.37 $\pm$ 0.08 & \textbf{88.01 $\pm$ 0.39} & \textbf{84.78 $\pm$ 0.29}
        \\
    \bottomrule
  \end{tabular}
\end{table*}


Table \ref{tab:results} shows the model performance and model-generated explanation quality for gender classification and scene recognition datasets. 
The results are obtained from 5 individual runs for every setting. 
The best results for each dataset are highlighted with boldface font and the second bests are underlined.
In general, our proposed framework variations, i.e., RES-G and RES-L, outperformed all other comparison methods in terms of both prediction accuracy as well as explainability on both datasets. 
Specifically, regarding prediction power, the RES-G with a pre-defined Gaussian transformation kernel as the imputation function achieved the best performance, outperforming the baseline DNN model by 4\% and 3\% on prediction accuracy on gender classification and scene recognition datasets, respectively.
In addition, the proposed RES framework enhanced the explainability of the backbone DNNs by a significant margin as compared with the baseline DNN model as well as other explanation supervision methods.
The proposed RES-L with learnable kernels as the imputation function achieved the biggest improvement on model explainability in terms of both IoU and F1 scores on both datasets, outperforming other comparison methods by 8\%-72\% and 16\%-36\% on IoU and explanation F1 scores, respectively.
The comparison methods GRADIA and HAICS also improved the model performance by leveraging the additional human attention labels, but are generally much less effective than the proposed RES framework. 
Those results demonstrated the effectiveness of the proposed framework on enhancing the model explainability robustly under noisy annotation labels, and consequently improved the model performance and prediction power on the prediction tasks.

\begin{figure*}
    \centering
    \includegraphics[width=0.9\textwidth]{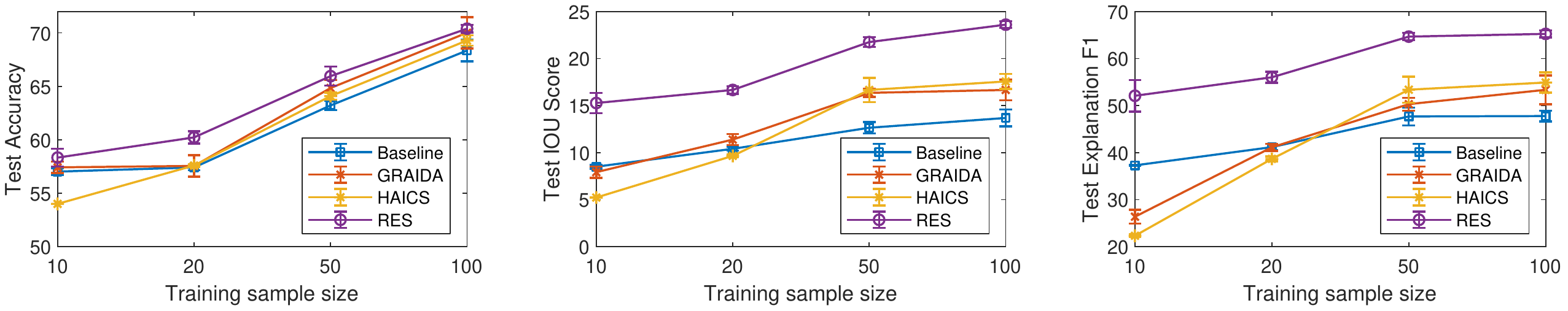}
    \vspace{-10pt}
    \caption{Model performance under different training sample size scenarios on gender classification dataset. The data point represents the mean value over 5 runs, and the error bar here corresponds to the standard deviation. (Left) The test prediction accuracy comparison. (Middle) The test IoU score comparison. (Right) The test explanation F1 score comparison.}
    \label{fig:trend}
\end{figure*}

Next, we further studied how the DNN models can benefit from the RES framework to gain a better generalization power under different training sample size scenarios. Specifically, we studied four training sample scenarios with training sample sizes of 10, 20, 50, and 100 on the Gender Classification Dataset. 
As shown in Figure~\ref{fig:trend}, we present the test prediction accuracy, IoU score, and explanation F1 score of each method under the four training sample size scenarios. 
The data point represents the mean value over 5 runs, and the error bar here corresponds to the standard deviation.
We can see that the proposed RES framework outperformed all other comparison methods by a significant margin under all scenarios studied, especially on boosting the explainability of the backbone DNNs as reflected by IoU and explanation F1 scores.
Specifically, RES was able to improve the model prediction accuracy by 2\% - 5\%, and boosted the quality of the model explanation by 60\%-80\% and 36\%-40\% in terms of IoU and explanation F1 scores, respectively.
Interestingly, we also observed degradation in model performance when applying GRADIA and HAICS when the sample size is extremely limited, such as in 10 and 20 training sample sizes scenarios. 
This could be due to the fact that GRADIA and HAICS simply treat the raw human annotation as clear data and thus suffer significantly from learning directly from the noisy labels and consequently prone to over-fitting badly. 
In contrast, with the robust learning objective, the proposed RES framework was able to cope with the noisy label pretty well even under a very limited sample size, and consequently boosted the model performance in terms of prediction power as well as explainability robustly in all scenarios studied.

\subsection{Qualitative Analysis of the Explanation}

\subsubsection{Case Studies}
\begin{figure*}
    \centering
    \includegraphics[width=0.9\textwidth]{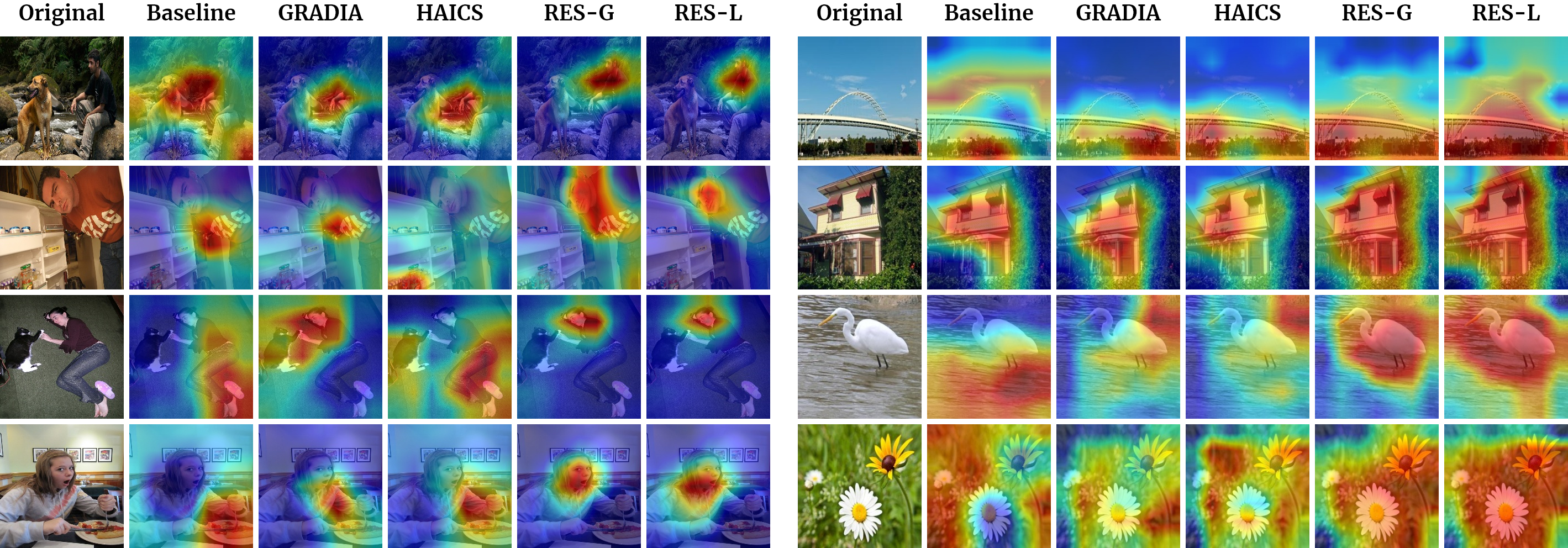}
    \caption{Selected explanation visualization results on gender classification dataset (left) and scene recognition dataset (right). The model-generated explanations are represented by the heatmaps overlaid on the original image samples, where more importance is given to the area with a warmer color.}
    \label{fig:exp_viz}
\end{figure*}
Here we provide some case studies about the model-generated explanation comparison for both gender classification and scene recognition datasets, as illustrated in Figure \ref{fig:exp_viz}. Here we present the model-generated explanations as the heatmaps overlaid on the original image samples, where more importance is given to the area with a warmer color.

\textbf{Gender Classification}:
As shown in the left four rows of Figure \ref{fig:exp_viz},  we studied two `male' class instances (top 2 rows) and two `female' class instances (bottom 2 rows). 
As can be seen, in general, the explanation generated by the proposed RES models can more accurately focus on the important areas (e.g., the human face areas) for identifying the gender of the person in the image. 
In contrast, both the baseline model as well as the two comparison methods failed to generate reasonable explanation, as the models' `attention' was distracted by some other objects presented in the images that are irrelevant to the gender classification task. 
For example, as shown in the first row on the left in Figure \ref{fig:exp_viz}, where both a dog and a person are presented in the image sample. The explanation generated by the baseline and comparison methods assigned importance to the areas in between the dog and the person, therefore, it could not focus properly on the person. On the other hand, both RES-G and RES-L learned to focus only on the person, more specifically on the facial area. 
Similar patterns could also be observed in the rest three rows on the left, demonstrating the powerful effect of the proposed RES framework on learning to generate more accurate explanations, and consequently enhance the explainability of the DNN models.

\textbf{Scene Recognition}:
For the scene recognition dataset, as shown in the right four rows in Figure \ref{fig:exp_viz}, we studied two instances of `urban' scene (top 2 rows) and two instances of `nature' scene (bottom 2 rows). 
Once again, we found that compared with the baseline model and other comparison methods, the explanations generated by RES models are more accurate and close to the ground truth for identifying whether the scene is taken from the urban areas or wild nature. 
For instance, as shown in the third row on the right in Figure \ref{fig:exp_viz}, the explanation generated by both the baseline and comparison methods focuses more on the water surface while RES focuses more on the wild animal itself. 
Similarly, as shown in the fourth row, the explanation generated by RES focuses more on the wildflowers than the grass-field background. 
Although in those situations the prediction can be correct for all the models studied, we argue that the model trained with the RES framework can be more robust and have a batter generalizability power to the downstream predictive tasks by learning to assign importance more accurately to the most distinguishable features/patterns presented in the data samples.

\subsubsection{Human Assessment}

\begin{figure}
    \centering
    \small
    \begin{tabular}{>{\raggedright\arraybackslash}p{32mm}|p{24mm}}
        \hline
        \textbf{Model Pairs} & \textbf{Perceived Quality (p-values)}
        \\ 
        \hline
        \textbf{Baseline vs. GRADIA} & \textbf{2.68e-03}\textsuperscript{\textdaggerdbl} 
        \\
        \textbf{Baseline vs. HAICS} & \textbf{2.33e-04}\textsuperscript{\textdaggerdbl\textdaggerdbl} 
        \\
        \textbf{Baseline vs. RES-G} & \textbf{4.98e-37}\textsuperscript{\textdaggerdbl\textdaggerdbl}
        \\
        \textbf{Baseline vs. RES-L} & \textbf{4.96e-28}\textsuperscript{\textdaggerdbl\textdaggerdbl}
        \\
        GRADIA vs. HAICS & 0.4980
        \\
        \textbf{GRADIA vs. RES-G} & \textbf{2.71e-22}\textsuperscript{\textdaggerdbl\textdaggerdbl}
        \\
        \textbf{GRADIA vs. RES-L} & \textbf{1.54e-15}\textsuperscript{\textdaggerdbl\textdaggerdbl}
        \\
        \textbf{HAICS vs. RES-G} & \textbf{1.67e-19}\textsuperscript{\textdaggerdbl\textdaggerdbl}
        \\
        \textbf{HAICS vs. RES-L} & \textbf{2.96e-13}\textsuperscript{\textdaggerdbl\textdaggerdbl}
        \\
        RES-G vs. RES-L & 0.0824
        \\
        \hline
    \end{tabular}
    \includegraphics[width=0.85\linewidth]{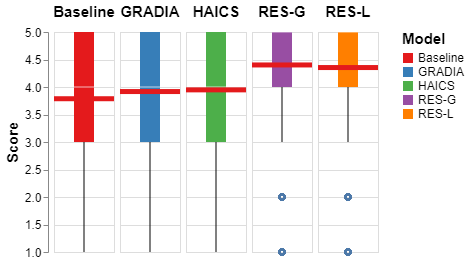}
    \vspace{-3mm}
    \caption{Top: results for pairwise comparison of five conditions. \textdagger: $p < 0.05$, \textdaggerdbl: $p < 0.01$, \textdaggerdbl\textdaggerdbl: $p < 0.001$. Bottom: Distributions of human users' perceived attention quality ratings. 5-level Likert scale is used (5: Excellent, 4: Good, 3: Fair, 2: Bad, 1: Inferior).}
    \label{fig:humanassessment}
\end{figure}

To evaluate the quality of explanations for the five comparison methods, we developed a web-based user interface (UI) where a human annotator can go over all the model-generated explanations and make qualitative evaluation on both datasets.
We distributed the model-generated explanations from the test set to three separate human annotators. 
We asked annotators to assess the perceived quality of explanations with the five-level Likert scale.
``5-Excellent'' when explanations show positive attention very clearly while don't contain negative attention at all,  and ``4-Good'' when positive attention is clearly presented with negligible negative attention.
``3-Fair'' meant that positive attention is partially seen while negative attention is clearly visible.
``2-Bad'' in case positive attention can be barely seen while negative can be found evidently.
``1-Inferior'' is assigned when a human annotator can only find negative attention.
After performing the Shapiro-Wilk normality test, we found participants' ratings don't follow a normal distribution.
Therefore, we applied Kruskal-Wallis H-test for identifying the differences between the five conditions.
The quality ratings of five models are significantly different, with a p-value of 7.82e-51 (< 0.05). For post-hoc pairwise comparisons using Dunn's test, all pairs are significantly different, with the exception of GRADIA vs. HAICS and RES-G vs. RES-L.
This means that the ranking among the five conditions is that RES-G (M = 4.40, SD = 0.91) and RES-L (M = 4.35, SD = 0.89) are rated notably higher than the rest, followed by GRADIA (M = 3.92, SD = 1.24) and HAICS (M = 3.95, SD = 1.23). The least performing condition was Baseline (M = 3.79, SD = 1.25). Specific pair-wise testing results and visual representation between conditions are shown in Figure ~\ref{fig:humanassessment}.

\begin{figure}
    \centering
    \includegraphics[width=0.9\linewidth]{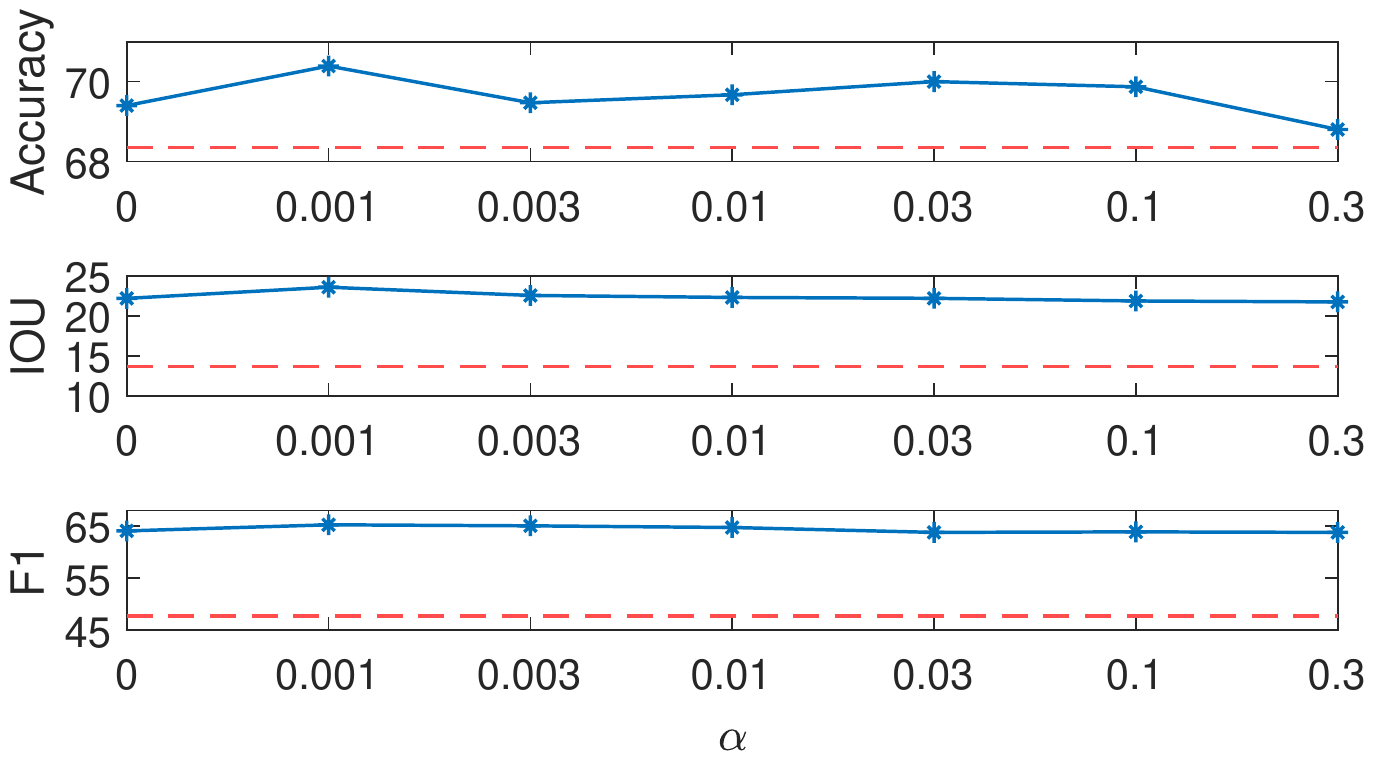}
    \vspace{-10pt}
    \caption{The sensitivity study of hyper-parameter $\alpha$ in RES framework (RES-L) on gender classification dataset. The red dashed lines represent the baseline model's performance. }
    \label{fig:eta}
\end{figure}

\subsection{Sensitivity Analysis of Hyper-parameter}
Here we further provide a sensitivity analysis of the hyper-parameter $\alpha$ introduced in the proposed RES framework, as shown in Equation \eqref{eq:exp_loss} which measures the tolerance level we give to the discrepancies between human annotation labels and the model explanation.
Figure \ref{fig:eta} shows the prediction accuracy, IoU, and explanation F1-score of the RES-L model for various values of $\alpha$ on the gender classification dataset. The scene recognition dataset follows a similar trend. The red dashed lines represent the baseline model's performance. 
In general, the model performance is not too sensitive to the value of $\alpha$ within the range studied, as all models outperformed the baseline model by a significant margin in terms of both prediction accuracy as well as explainability. 
As we developed our models based on the accuracy of the validation set, we indeed observed a concave curvature on test accuracy, peaking at a $\alpha$ value between 0.001 and 0.1. 
While the specific best value of $\alpha$ can vary depending on the dataset as well as the degree of nosiness of the human annotation labels (such as the granularity of the annotation), in general, the proposed framework can perform well when $\alpha$ is relatively small (e.g., less than 0.1).

\section{Conclusion}
This paper proposes a generic framework for visual explanation supervision by developing a novel explanation model objective that can handle the noisy human annotation labels as the supervision signal with a theoretical justification of the benefit to model generalizability.
Extensive experiments on two real-world image datasets demonstrate the effectiveness of the proposed framework on enhancing both the reasonability of the explanation as well as the performance of the backbone DNNs model. Although the additional data of human explanation labels may not be easily accessible, our studies have demonstrated the effectiveness of the proposed RES framework under a quite limited amount of training samples, which could benefit application domains where data samples are limited and hard to acquire, yet both model performance as well as the explainability are on-demand, such as in medical domains.
Furthermore, designing effective semi-supervised or weakly-supervised explanation supervision frameworks can be promising future directions to further overcome this limitation.  

\begin{acks}
This work was supported by the NSF Grant No. 1755850, No. 1841520, No. 2007716, No. 2007976, No. 1942594, No. 1907805, NSF Future of Work grant No. 2026513, a Jeffress Memorial Trust Award, Amazon Research Award, NVIDIA GPU
Grant, and Design Knowledge Company (subcontract number: 10827.002.120.04).
\end{acks}

%
\bibliographystyle{ACM-Reference-Format}
\bibliography{sample-base}

\clearpage
\appendix
\section{Appendix}

\subsection{Proof of Theorem \ref{th:1}}
\label{ap:th1}
\begin{proof}

Suppose $f_{\psi}$ is a $\delta$-minimizer of $\mathcal{L}$ with $\psi\in\Theta$. From Assumption~\autoref{ass:1}, we know that there exists a neural network $f_{\tau}$ such that 
\begin{equation}
    \Vert f_{\tau} - f_{\theta^{*}}\Vert^2 \coloneqq \mathbb{E}\left[\vert f_{\tau} - f_{\theta^{*}}\vert^2 + \vert \nabla f_{\tau} - \nabla f_{\theta^{*}}\vert^2\right] \leq C_{1}^2 \frac{\Vert \theta^{*}\Vert^2}{m^{\gamma}}
\end{equation}
Then, we have
\begin{equation}
    \begin{split}
    \mathcal{L}\!(f_{\psi})\!-\!\mathcal{L}\!(f_{\theta^{*}}\!) \!&\leq\!\mathcal{L}(f_{\tau}) - \mathcal{L}(f_{\theta^{*}}) + 
    \delta \\
    &\!\leq\! L_1 \mathbb{E}\left[\vert f_{\tau}(x)\!-\! f_{\theta^{*}}(x)\right] +\!L_2\mathbb{E}\left[\vert\nabla f_{\tau}(x)\!-\!\nabla f_{\theta^{*}}(x) \right]\!+\!\delta \\
    &\!\leq\! (L_1 + L_2) \frac{C_1\Vert \theta^{*}\Vert}{m^{\gamma/2}} + \delta
    \end{split}
\end{equation}

From Assumption~\autoref{ass:2}, given $0<\epsilon<1$, we have
\begin{equation}
    P(\vert \mathcal{L}(f_{\theta}) - \hat{\mathcal{L}}(f_{\theta})\vert \leq \frac{C_2 (V,m,\epsilon)}{\sqrt{N}}) \geq 1-\epsilon,\quad\forall\;\theta\in\Theta
\end{equation}
Then,
\begin{equation}
    \begin{split}
        \mathcal{L}(f_{\hat{\theta}}) - \mathcal{L}(f_{\theta^{*}}) &\leq \hat{\mathcal{L}}(f_{\hat{\theta}}) - \mathcal{L}(f_{\theta^{*}}) + \frac{C_2 (V,m,\epsilon)}{\sqrt{N}} \\
        & \leq \hat{\mathcal{L}}(f_{\psi}) - \mathcal{L}(f_{\theta^{*}}) + \frac{C_2 (V,m,\epsilon)}{\sqrt{N}} + \delta \\
        &\leq \mathcal{L}(f_{\psi}) - \mathcal{L}(f_{\theta^{*}}) + \frac{C_2 (V,m,\epsilon)}{\sqrt{N}} + \delta \\
        &\leq (L_1 + L_2) \frac{C_1\Vert \theta^{*}\Vert}{m^{\gamma/2}} + \frac{2C_2 (V,m,\epsilon)}{\sqrt{N}} + 2\delta
    \end{split}
\end{equation}
\end{proof}

\subsection{Proof of Lemma 1}
\label{ap:lem1}
\begin{proof}
Since 
\begin{equation}
tanh(x) = \frac{e^x - e^{-x}}{e^x + e^{-x}} = \frac{1 - e^{-2x}}{1 + e^{-2x}}
\end{equation}
where the last equality follows by multiplying by $\frac{e^{-x}}{e^{-x}} = 1$. And since: $\lim_{x\to\infty} 1 - e^{-2x} = 1$, and $\lim_{x\to\infty} 1 + e^{-2x} = 1$, we have
\begin{equation}
\lim_{x\to\infty} tanh(x) = 1
\end{equation}
Similarly, we also have
\begin{equation}
\lim_{x\to-\infty} tanh(x) = \lim_{x\to-\infty} \frac{e^{2x} - 1}{e^{2x} + 1} = -1
\end{equation}
Thus we have
\begin{equation}
\lim_{\gamma\to\infty}tanh(\gamma(M^{(i)}-a)) = \left\{
        \begin{array}{ll}
            1  & \quad M^{(i)}>a \\
            -1 & \quad M^{(i)}<a \\
        \end{array}
    \right.
\end{equation}
Thus we have the equivalency of Equation \eqref{eq:tanh} and Equation \eqref{eq:exp_loss} when $\gamma \rightarrow \infty$.
\end{proof}


\subsection{Human Annotation and Evaluation UI demonstration}
\label{ap:UI}

\begin{figure*}[t]
\centering
\begin{subfigure}[b]{0.95\textwidth}
  \includegraphics[width=1\linewidth]{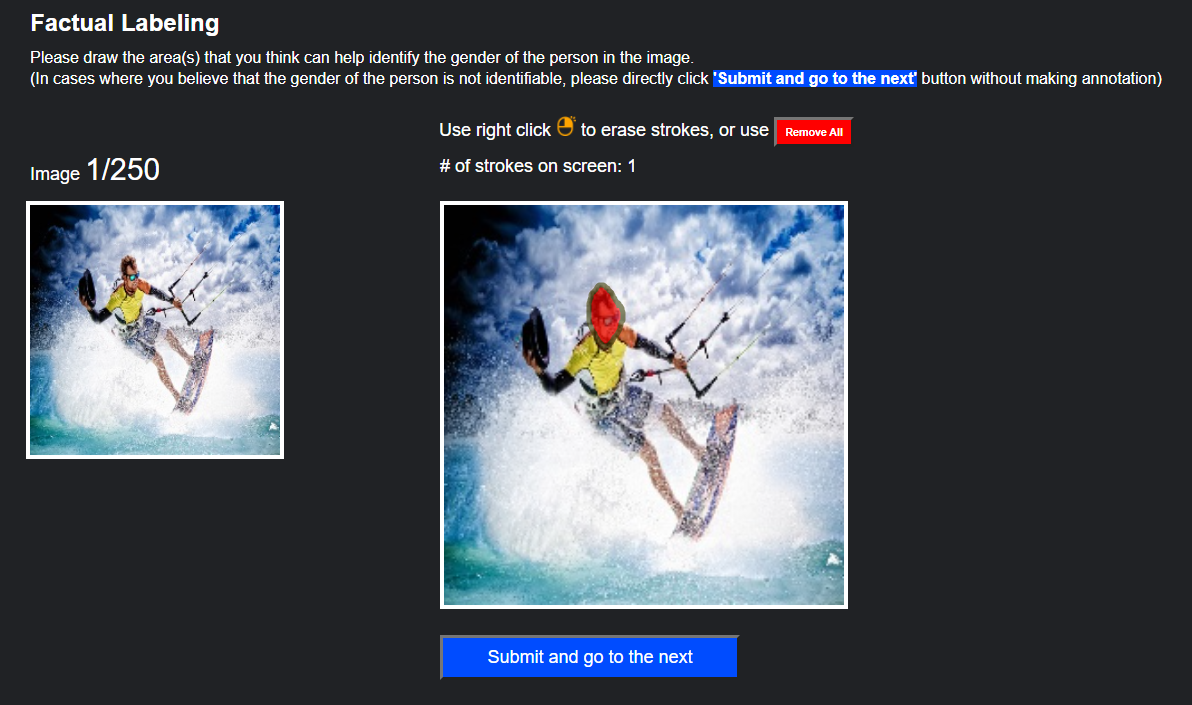}
  \caption{}
  \label{fig:ui_draw}
\end{subfigure}

\begin{subfigure}[b]{\textwidth}
  \includegraphics[width=1\linewidth]{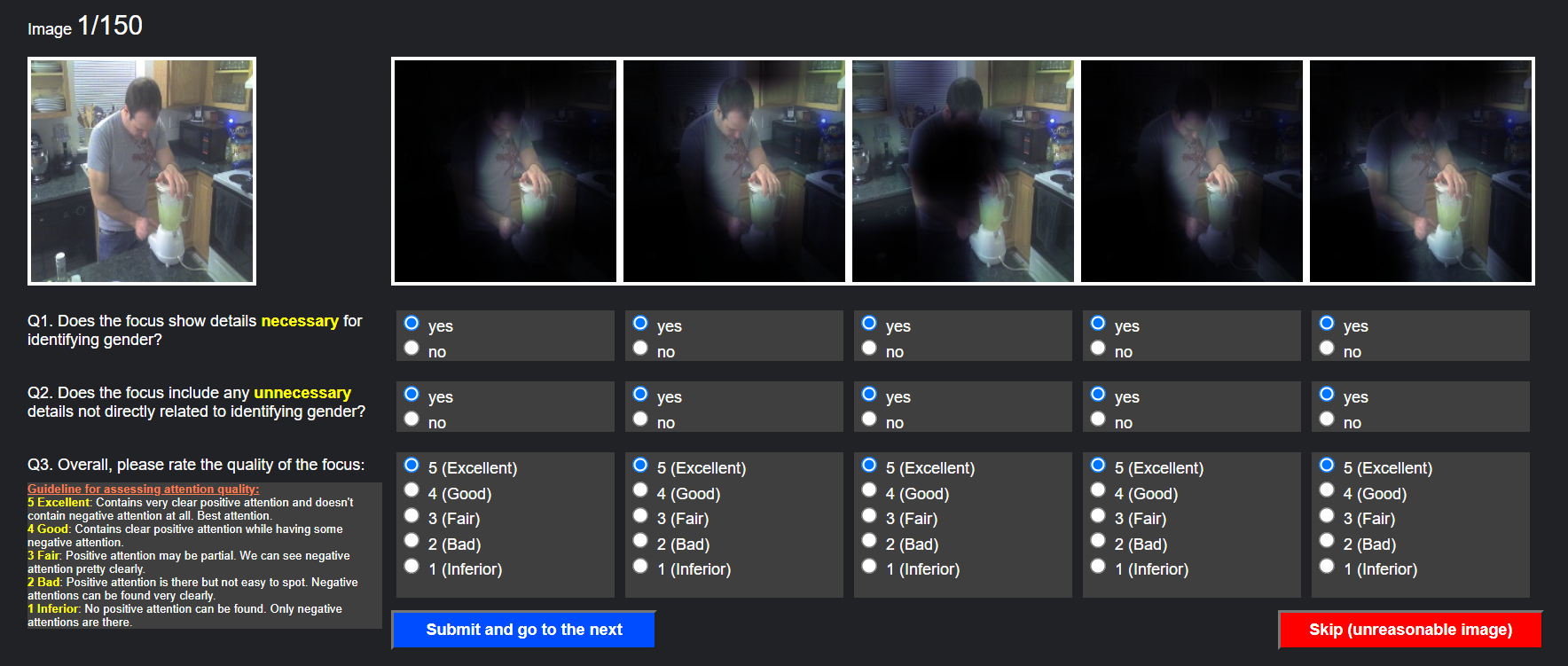}
  \caption{}
  \label{fig:ui_rate}
\end{subfigure}
\vspace{-10pt}
\caption{The screenshots illustrating the two UIs for human annotation and evaluation. (a) The interface for attention annotation where users can draw on the image and generate a binary matrix of the focus area used for improving model explanation quality. (b) The interface for attention quality assessment where 5 model-generated explanations are displayed in random order. Users will answer three questions for each explanation.}
\label{fig:ui}
\end{figure*}

Figure \ref{fig:ui} (a) is the interface used to collect attention annotation on the areas people think are relevant to the classification task. For example, for the gender dataset annotation, users first determine whether they can identify the person's gender in the image, then draw the areas that help them for the gender classification. In the back-end, the coordinates of highlighted areas are converted into a binary map, preparing for the modeling step.

Figure \ref{fig:ui} (b) is the interface for human assessment on the model-generated explanations. For each image annotation, 5 explanations were presented in random order with 3 questions (Q1 and Q2 are true/false questions, Q3 is a 5-point Likert scale rating question) asked for each explanation. Question 1 asks if the focus on the explanation shows details necessary for identifying the target label (i.e., labels in gender classification or scene recognition), and question 2 asks for the presence of unnecessary details on the image for identifying the target. Question 3 is our main focus of the attention quality assessment, where annotators give 1 to 5 ratings to each model explanation.



\subsection{Efficient Adaptive Threshold Searching Algorithm}
\label{ap:a}
\begin{algorithm}
\small
\caption{Adaptive Threshold Searching Algorithm}
\begin{varwidth}{\linewidth}
\begin{algorithmic}[1]
  \REQUIRE $M, F, C$
  \ENSURE solution $a$
  \STATE initialize: $a=0, act=0, v=0, vct=0, i=0, j=0$
  \STATE {$ge = \{M[find(C>0)]\}$ \textit{\% find the set of greater or equal to inequality constraints}}
  \STATE {$l = \{M[find(F>0)]\}$ \textit{\% find the set of less to inequality constraints}}
  \STATE {$ges = \text{Sort}(ge, \text{`ascend'})$}
  \STATE {$ls = \text{Sort}(l, \text{`descend'})$}
    \FOR{$i<|ges|$}                    
        \STATE {$v=ges[i]$}
        \STATE {$vct=i+1+\text{BinarySearch}(v, ls)$}
        \IF {$vct > act$}
            \STATE {$a = v$}
            \STATE {$act = vct$}
        \ENDIF
        \STATE {$i=i+1$}
    \ENDFOR
    \FOR{$j<|ls|$}                    
        \STATE {$v=ls[i]$}
        \STATE {$vct=j+1+\text{BinarySearch}(v, ges)$}
        \IF {$vct > act$}
            \STATE {$a = v$}
            \STATE {$act = vct$}
        \ENDIF
        \STATE {$j=j+1$}
    \ENDFOR
\end{algorithmic}
\end{varwidth}
\end{algorithm}

\subsection{Detailed Implementation of the Learnable Imputation Layers}
\label{ap:imp}
For the learnable imputation function, we studied both a shallow implementation as well as a deep implementation, as shown in detail below:

\textbf{Shallow Implementation}: We apply one layer of convolution operation to process the raw human annotation label, with a $64\times64$ convolution kernel with a padding size of 16 and a stride of 32. 

\textbf{Deep Implementation}: We apply five layers of convolution operations to process the raw human annotation label, with $7\times7$, $3\times3$,  $3\times3$,  $3\times3$, and $3\times3$ convolution kernel with a padding size of 3 on the first layer and 1 for the rest layer, and a stride 2 for all layers.

We choose the Shallow implementation for the RES-L model as it achieves better performance on the validation set. The reason why the deep version gets inferior performance could be due to the training sample size studied in this work is too small.


\end{document}